# Improving the Efficiency of Approximate Inference for Probabilistic Logical Models by means of Program Specialization

Daan Fierens

Katholieke Universiteit Leuven, Department of Computer Science,
Celestijnenlaan 200A, 3001 Heverlee, Belgium
`Daan.Fierens@cs.kuleuven.be`

**Abstract.** We consider the task of performing probabilistic inference with probabilistic logical models. Many algorithms for approximate inference with such models are based on sampling. From a logic programming perspective, sampling boils down to repeatedly calling the same queries on a knowledge base composed of a static and a dynamic part. The larger the static part, the more redundancy there is in the repeated calls. This is problematic since inefficient sampling yields poor approximations.
We show how to apply logic program specialization to make sampling-based inference more efficient. We develop an algorithm that specializes the query-predicates with respect to the static part of the knowledge base. In experiments on real-world data we obtain speedups of up to an order of magnitude, and these speedups grow with the data-size.

## 1 Introduction

In the field of artificial intelligence there is a large interest in *probabilistic logical models*, namely probabilistic extensions of logic programs as well as first-order logical extensions of probabilistic models such as Bayesian networks [3,5,11]. *Probabilistic inference* with such a model is the task of answering various questions about the probability distribution specified by the model, usually conditioned on certain observations (the *evidence*). While a variety of inference algorithms is being used, many of them are based on *sampling* [1,2,9,13]. The idea is to draw samples from the considered probability distribution conditioned on the evidence, and use these samples to compute an *approximate* answer to the inference questions of interest. It is important that the process of sampling is efficient because the more samples can be drawn per time-unit, the more accurate the answer will be (i.e., the closer to the correct answer).

In this paper we use logic programs to represent probabilistic logical models and we use Prolog as the programming environment in which we implement sampling-based inference. From an implementation perspective, sampling boils down to repeatedly calling the same queries on a knowledge base composed of a static part (the evidence) and a highly dynamic part that changes at runtime because of the sampling. The more evidence, the larger the static part of the knowledge base, so the more redundancy there is in these repeated calls. Since it is important that the sampling process is efficient, this redundancy needs to be reduced as much as possible. In this paper we show how to do this by applying *logic program specialization*: we specialize the definitions of the query-predicates with respect to the static part of the knowledge base. While a lot of work about logic program specialization is about exploiting static information about the input arguments of queries (partial deduction [6]), we instead exploit static information about the knowledge base on which the queries are called.

The *contributions of this paper* are three-fold. First, we show how to represent the considered models and implement sampling-based inference algorithms for them in Prolog (Sections 3 and 4). Second, we argue that logic program specialization can be used to make sampling in the presence of evidence more efficient and we develop an appropriate specialization algorithm (Section 5). Third, we evaluate our approach with experiments on real-world data. Our results show that specialization can significantly speed up the sampling process and that the speedups grow with the data-size (Section 6).

This paper extends an earlier paper [4]. First, this paper contains the entire specialization algorithm, whereas the earlier paper describes only the outer loops of the algorithm but not the inner loops. Second, in this paper we discuss sampling-based inference algorithms in general rather than only one specific algorithm. Third, this paper contains more experimental results (with varying data-sizes). Fourth, this paper is more complete on various aspects (support for meta predicates, description of the experimental setup, etc.).

Our approach is applicable to all kinds of probabilistic logical models and programs. In this paper we focus on models that are first-order logical or "relational" extensions of Bayesian networks [3,5]. Concretely, we use the general framework of *parameterized Bayesian networks* [11]. Before introducing this framework, we first give some background on probability theory and Bayesian networks.

## 2 Background

**Probability Theory.** In probability theory [9] one models the world in terms of *random variables (RVs)*. Each state of the world corresponds to a joint state of all considered RVs. We use upper case letters to denote single RVs and bold-face upper case letters to denote sets of RVs. We refer to the set of possible states/values of an RV $X$ as the *range* of $X$, denoted $range(X)$. We consider only RVs with a finite range (i.e., discrete RVs). A *probability distribution* for an RV $X$ is a function that maps each $x \in range(X)$ to a number $P(x) \in [0,1]$ such that $\sum_{x \in range(X)} P(x) = 1$. A *conditional probability distribution (CPD)* for an RV $X$ conditioned on a set of other RVs $\mathbf{Y}$ is a function that maps each possible joint state of $\mathbf{Y}$ to a probability distribution for $X$.

**Bayesian Networks.** A *Bayesian network* [9] for a set of RVs **X** is a set of CPDs: for each $X \in \mathbf{X}$ there is one CPD for $X$ conditioned on a (possibly empty) set of RVs called the *parents* of $X$. Intuitively, the CPD for $X$ specifies the direct probabilistic influence of $X$'s parents on $X$. The probability distribution for $X$ conditioned on its parents $\mathbf{pa}(X)$, as determined by the CPD for $X$, is denoted $P(X \mid \mathbf{pa}(X))$.

Semantically, a Bayesian network represents a joint probability distribution $P(\mathbf{X})$ on the set of possible joint states of **X**. Concretely, $P(\mathbf{X})$ is the product of the CPDs in the Bayesian network: $P(\mathbf{X}) = \prod_{X \in \mathbf{X}} P(X \mid \mathbf{pa}(X))$. It can be shown that $P(\mathbf{X})$ is a proper probability distribution if the parent relation between the RVs (as visualized by a directed graph) is acyclic.

## 3 Parameterized Bayesian Networks

Bayesian networks essentially use a propositional representation. Several ways of lifting them to a first-order representation have been proposed [3, Ch.6,7,13] [5,13]. There also exist several probabilistic extensions of logic programming, such as PRISM, Independent Choice Logic and ProbLog [3, Ch.5,8]. These different kinds of probabilistic logical models essentially all serve the same purpose. In this paper we focus on the Bayesian network approach. While there are many different frameworks for first-order logical (or "relational") extensions of Bayesian networks, we focus on *parameterized Bayesian networks* [11] since they offer a simple yet adequate representation.

Parameterized Bayesian networks use the concept of a *parameterized RV* or *parRV*. A parRV has one or more typed *parameters*. Each parameter type has a *population*. When each parameter in a parRV is instantiated/grounded to a particular element of its population we obtain a regular RV. A parameterized Bayesian network is simply a set of *parameterized CPDs*, namely one for each parRV. Rather than providing a formal discussion of parameterized Bayesian networks we show how they can be represented as logic programs in Prolog.

**Representation.** To deal with parRVs we associate to each of them a unique predicate. For a parRV with $n$ parameters we use a $(n+1)$-ary predicate: the first $n$ arguments are the parameters, the last argument is the state of the RV. We call these predicates *state predicates*.

Each parRV has one corresponding parameterized CPD. To deal with parameterized CPDs we again associate to each of them a unique predicate, the last argument is now a probability distribution on the range of the corresponding parRV. We call these predicates *CPD-predicates*. We assume that each CPD-predicate is defined by a decision list. A *decision list* is an ordered set of rules such that there is always at least one rule that applies, and of all rules that apply only the first one fires (in Prolog this is achieved by putting a cut at the end of each body and having a last clause with *true* as the body).

*Example 1.* Consider a university domain with students and courses. Suppose that we use the following parRVs: level (with a parameter from the population of courses), IQ and graduates (each with a student parameter) and grade (with a student parameter and a course parameter). To represent the state of the RVs we use the state predicates *level*/2, *iq*/2, *graduates*/2 and *grade*/3. For instance, the meaning of the predicate *level*/2 is that the atom *level*(*C*, *L*) is true if the RV *level* for the course *C* is in state *L*.

To represent the parameterized CPDs we use CPD-predicates *cpd_level*/2, *cpd_iq*/2, *cpd_grade*/3 and *cpd_graduates*/2. Suppose that the parRV *level* does not have parents, then its parameterized CPD could for instance look as follows.

```
cpd_level(_C,[intro:0.4,advanced:0.6]).
```

We use lists like `[intro:0.4,advanced:0.6]` to represent probability distributions. The other parameterized CPDs could for instance be defined as follows.

```
cpd_iq(_S,[high:0.5,low:0.5]).

cpd_grade(S,C,[a:0.7,b:0.2,c:0.1]) :- iq(S,high),level(C,intro),!.
cpd_grade(S,C,[a:0.2,b:0.2,c:0.6]) :- iq(S,low),level(C,advanced),!.
cpd_grade(_S,_C,[a:0.3,b:0.4,c:0.3]).

cpd_graduates(S,[yes:0.2,no:0.8]) :- grade(S,_C,c), !.
cpd_graduates(S,[yes:0.5,no:0.5]) :- findall(C,grade(S,C,a),L),
                                    length(L,N), N<2, !.
cpd_graduates(_S,[yes:0.9,no:0.1]).
```
 □

In the bodies of the clauses defining the CPD-predicates we allow the use of state predicates (like *iq*/2 and *level*/2 in the clauses for *cpd_grade*/3), but not of CPD-predicates. Note that CPD-predicates can be defined by non-ground facts (e.g. *cpd_level*/2). This does not cause problems since CPD-predicates are always called with all arguments except the last instantiated (see Section 4.3).

**Semantics.** Given the population for each parameter type (e.g. the set of courses and the set of students), the set of all RVs **X** is defined as the set of RVs that can be obtained by grounding all parameters in a parRV with respect to these populations. A parameterized Bayesian network determines a joint probability distribution on the set of possible joint states of **X**, in a similar way as a regular Bayesian network does: $P(\mathbf{X}) = \prod_{X \in \mathbf{X}} P(X \mid \mathbf{pa}(X))$, where $P(X \mid \mathbf{pa}(X))$ denotes the probability distribution for $X$ as determined by the corresponding parameterized CPD.

## 4 Sampling-based Probabilistic Inference

Typically the state of a subset of the RVs **X** is known. This information is called the *evidence*. We call an RV *observed* if it is in the evidence and *unobserved* otherwise. *Probabilistic inference* is the task of answering questions about the probability distribution $P(\mathbf{X})$ conditioned on the evidence. The most common inference task is to compute marginal probabilities of unobserved RVs. A *marginal probability* is the probability that a particular RV is in a particular state. For instance, given the grades of all students for all courses (the evidence), we may want to find for each student the probability that he has a high IQ. Such probabilities can in theory be computed by performing arithmetic on the probabilities specified in the parameterized CPDs, but for real-world population sizes this is intractable (inference with Bayesian networks is NP-hard [9]). Hence, one often uses *approximate inference* instead.

### 4.1 Sampling-based Approximate Inference

An important class of approximate probabilistic inference algorithms are *sampling-based* ('Monte Carlo') algorithms such as rejection sampling, importance sampling, Gibbs sampling, MCMC, and many variants [1,2,13]. All these algorithms draw samples from the probability distribution $P(\mathbf{X})$ conditioned on the evidence. A *sample* is an assignment of a value to each relevant RV. To effectively condition on the evidence, only samples that are consistent with the evidence are considered (a sample is consistent if it assigns to each observed RV its known value). Given $N$ such samples, we can construct an approximate answer to the inference questions. For marginal probabilities this is straightforward. For instance, the marginal probability that student $s1$ has a high IQ conditioned on the evidence is estimated as the number of samples in which the RV $iq$ for $s1$ has value 'high', divided by $N$. For all common sampling algorithms, it holds (under some restrictions) that: the higher $N$, the more accurate the estimates (i.e. the closer they are to the correct value, with convergence for $N \to \infty$) [1].

Sampling-based inference is often used by giving the sampling process a fixed time to run before computing the estimates. The less time it takes to draw a sample, the more samples can be drawn in the given time, so the higher the accuracy of the estimates. In other words: *any gain in efficiency of the sampling process can lead to a gain in accuracy*. Hence it is crucial to implement the sampling process as efficiently as possible. Before we show how program specialization can help with this, we explain how sampling-based algorithms typically work.

### 4.2 Generic Structure of Sampling-based Inference Algorithms

A sample is an assignment of a state (value) to each RV. Hence the main data structure used in sampling algorithms is a structure that stores the 'current state' of all RVs. In essence, sampling is a process that continuously modifies this data structure in a stochastic way.

The generic structure common to many sampling-based inference algorithms is shown in Figure 1. As explained, we are only interested in samples that are consistent with the evidence. Hence we start by initializing all observed RVs to their known value (line 1 and 2 in Figure 1). These RVs stay in this state during the entire sampling process. Next we start drawing samples (line 3). To draw one

```
   // initialize observed RVs according to the evidence:
 1 for each observed RV O
 2     set O to its known value
   // draw samples:
 3 repeat until enough samples
 4     for each unobserved RV U
           // (re)sample U:
 5         compute $P_{sample}(U)$
 6         sample u from $P_{sample}(U)$
 7         set U to u
```

**Fig. 1.** Generic structure of sampling-based inference algorithms.

sample, a typical approach is to visit each unobserved RV $U$ and '(re)sample' it. This is done as follows.

1) We compute the distribution $P_{sample}(U)$ that we will sample from (line 5). Which distribution is used depends on the considered sampling algorithm (see below). As an example, for the RV that indicates the grade of a student for a course, we might for instance get the distribution `[a:0.7,b:0.2,c:0.1]`.
2) We sample a state from this distribution (line 6). In our example, there is for instance a 70% probability that the state will be the value 'a'.
3) We set the RV $U$ to this new state by modifying the data structure (line 7).

This resampling step needs to be performed for each unobserved RV and for each sample. It is common to have thousands of RVs and to draw tens of thousands of samples (see Section 6.1). This means that the resampling step needs to be executed several millions of times or more.

Perhaps the most essential difference between various sampling-based inference algorithms is how they determine the distribution $P_{sample}(U)$.

- Simple algorithms such as forward sampling and rejection sampling [2] define $P_{sample}(U)$ as the distribution on $U$ conditioned on the current state of $U$'s parents. This distribution can be found by *applying the CPD for U*. For instance, suppose that $U$ is the RV that indicates the grade of student $s1$ for course $c1$. According to the CPD for $U$ (given in Example 1, Section 3) the distribution on $U$ depends on the current state of the IQ of $s1$ and the level of $c1$. Suppose that the IQ is 'high' and the level is 'intro'. The CPD then states that the distribution on $U$ is `[a:0.7,b:0.2,c:0.1]`. This distribution is then used as the distribution $P_{sample}(U)$. Since we represent each CPD as a decision list it is guaranteed that this always returns exactly one probability distribution (as required by the semantics of parameterized Bayesian networks).
- More advanced algorithms such as Gibbs sampling [1,13] and other MCMC variants [2] use a more complex definition of $P_{sample}(U)$. Computing this distribution typically requires applying multiple CPDs and performing some simple arithmetic to combine the results into the distribution $P_{sample}(U)$ [1,4].

To summarize, resampling an unobserved RV $U$ requires the following operations:

1) apply one or more CPDs (and perform some arithmetic on the results in order to calculate the distribution $P_{sample}(U)$),
2) sample a new state from this distribution,
3) set $U$ to this new state.

Operation 2 is straightforward. In the next section we explain how we implement operations 1 and 3 in our Prolog implementation for parameterized Bayesian networks.

### 4.3 Sampling: The Prolog Implementation Perspective

**Data structure.** As explained we need to store the current state of all RVs. This is done by keeping a knowledge base with facts defining the state predicates. For instance, a fact $grade(s1, c1, b)$ indicates that student $s1$ currently has grade 'b' for course $c1$. We call this knowledge base (KB) the *state KB*.

**Operations.** We need to perform the following operations on the state KB.

- **Setting an RV.**
  Setting an RV to a given state is done by modifying the corresponding fact in the state KB. For instance, the fact $grade(s1, c1, b)$ is turned into $grade(s1, c1, a)$. For efficiency reasons we do not use assert and retract to update the state KB but instead internally modify the last argument of the fact directly.
- **Applying a CPD.**
  Applying the CPD for an RV is done by calling the associated CPD-predicate on the state KB. For instance, for the RV that indicates the grade of student $s1$ for course $c1$, applying the CPD is done by calling $cpd\_grade(s1, c1, Distr)$ on the state KB. This returns the desired distribution $Distr$.
  We refer to such a query as a *CPD-query*. To be precise, a CPD-query is any query built of a CPD-predicate (e.g. $cpd\_grade$) that has its last argument uninstantiated (e.g. $Distr$) and all other arguments instantiated to elements of the proper populations (e.g. student $s1$ and course $c1$).

From a Prolog implementation perspective, resampling an RV $U$ (line 5 to 7 in the algorithm of Figure 1) requires the following operations:

1) call one or more CPD-queries on the state KB and perform some arithmetic on the results in order to calculate the distribution $P_{sample}(U)$,
2) sample a new state from this distribution,
3) set $U$ to this new state by modifying the state KB.

In experiments (with the Gibbs sampling algorithm for parameterized Bayesian networks [4]) we found that there is one operation that is clearly the computational bottleneck, namely calling CPD-queries.

There is a fixed set of CPD-queries, namely one CPD-query for each RV. Each of these CPD-queries is typically called many times during the entire sampling process (because we draw many samples). The state KB on which the CPD-queries are called has a static part (the state of the evidence RVs never changes) and a highly dynamic part (the state of the unobserved RVs changes continuously because of the sampling). The dynamic part implies that repeated calls of the same CPD-query can in general produce different answers. The static part, on the other hand, causes some redundancy in repeated calls of the same CPD-query: any (sub)computations that are local to the static part of the state KB will be performed over and over again in exactly the same way. Recall that for the sake of accuracy we want the sampling process to be as efficient as possible. Hence this redundancy needs to be removed.

The more evidence, the larger the static part of the state KB, so the larger the redundancy. In many applications there is a large amount of evidence. A common inference scenario for which this holds is *classification*: all RVs associated to one parRV (the class) are unobserved and need to be predicted based on observations of all the other RVs. For instance, we can predict the class of web pages based on observations of their textual content and hyperlinks [12]. Another common scenario is dealing with *missing data*: when we have data in which a minority of all entries are missing (due to measurement or storage errors), we can use probabilistic inference to fill in the missing entries based on the observed ones. This is for instance often done for machine learning from incomplete data [9, Ch.8.3], [8]. In both scenarios there are typically more observed RVs than unobserved ones.

To summarize, we often have a large amount of evidence and this causes redundancy in the sampling process. In the next section we show how this redundancy can be removed and sampling be made more efficient.

## 5  Applying Logic Program Specialization to CPDs

A solution to the above redundancy is to *specialize the definitions of the CPD-predicates with respect to the static part of the state KB (the evidence)*. Recall that each CPD-predicate is defined by a set of Prolog clauses that form a decision list, and that the bodies of these clauses refer to the state predicates. The evidence is a partial interpretation of the state predicates. Our specialization approach is a source-to-source transformation that takes as input the decision lists for all the CPD-predicates plus the evidence, and returns as output a specialized version of the decision lists.

The decision lists are used during inference in only one way, namely by calling CPD-queries, and there is a fixed set of possible CPD-queries. For each CPD-query, our specialized decision lists return the same answer as the original

```
procedure SPECIALIZE(U, O, o)                    procedure SPEC_DECISION_LIST(D, q, U, O, o)
1  for each CPD-predicate p                      1  if D is non-empty
2     let D be the decision list for p           2     let C be the first clause in D
3     Q=GET_CPD_QUERIES(p, U ∪ O)                       and D_rest be the other clauses in D
4     for each q ∈ Q                             3     C_q = GROUND_HEAD(C, q)
5        SPEC_DECISION_LIST(D, q, U, O, o)       4     let H_q be the head and B_q the body of C_q
                                                 5     B_spec=SPECIALIZE_BODY(B_q, U, O, o)
function SPECIALIZE_BODY(B, U, O, o)             6     if B_spec = true
1  B_1 = GROUND_BODY(B, U ∪ O)                   7        ASSERT_FACT(H_q)
2  B_2 = SPECIALIZE_LITERALS(B_1, U, O, o)       8     else if B_spec = false
3  B_3 = SIMPLIFY_BODY(B_2)                      9        SPEC_DECISION_LIST(D_rest, q, U, O, o)
4  if B_3 is identical to B_1                    10    else
5     return B                                   11       ASSERT_CLAUSE(H_q, B_spec)
6  else return B_3                               12       SPEC_DECISION_LIST(D_rest, q, U, O, o)
```

**Fig. 2.** The outer loops of the specialization algorithm.

decision lists. This ensures that sampling produces exactly the same sequence of samples with specialization as without (but in a more efficient way).

There is a lot of work on specialization, or more generally transformation, of logic programs that has the same end-goal as our work, namely to derive from a given program an 'equivalent' but more efficient program [10]. However, we are not aware of work that considers the same setting as we do, namely calling a fixed set of queries on a knowledge base with a static and a dynamic part, and specializing with respect to the static part. In particular, this setting makes our work different from the work on *partial deduction* for logic programs [6,7]. The essence of partial deduction is to exploit staticness of a subset of the arguments of queries. In our setting, we instead exploit staticness of part of the knowledge base on which the queries are called. Hence, existing systems for partial deduction (see e.g. Leuschel et al. [7]) are, as far as we see, not optimal for our setting.

### 5.1 Outer Loops of the Specialization Algorithm

We specialize the CPD-predicates with respect to the evidence. Because the evidence is at the ground level but the CPD-predicates are defined at the non-ground level, we first (partially) ground these definitions before we specialize them. The specialization algorithm is shown in Figure 2. **U** denotes the set of unobserved RVs, **O** the set of observed RVs and **o** their observed values (the evidence).

The outer-loop of the algorithm (line 1 of the SPECIALIZE procedure) is over all CPD-predicates: we specialize each CPD-predicate $p$ in turn. To do so, we collect the set of all possible CPD-queries for $p$ (line 3) and we loop over this set: for each CPD-query $q$ we apply the SPEC_DECISION_LIST procedure (line 5). We explain the latter with an example.

*Example 2.* Let $p$ be *cpd_graduates*/2, let the decision list $D$ that defines $p$ be the same as given earlier (Example 1, Section 3), and let the CPD-query $q$ be

$cpd\_graduates(s1, Distr)$. The SPEC_DECISION_LIST procedure starts by processing the first clause in $D$, namely:

```
cpd_graduates(S,[yes:0.2,no:0.8]) :- grade(S,_C,c), !.
```

First we ground the head variables of this clause with respect to $q$ (line 3 of the procedure SPEC_DECISION_LIST), yielding the clause $C_q$:

```
cpd_graduates(s1,[yes:0.2,no:0.8]) :- grade(s1,_C,c), !.
```

Next, we apply the function SPECIALIZE_BODY to the body of $C_q$ (line 5), yielding the new body $B_{spec}$. There are three possible cases.

- If $B_{spec}$ equals *true*, we assert a fact `cpd_graduates(s1,[yes:0.2,no:0.8])` (line 7). We can then discard the remaining clauses in $D$ with respect to $q$ (these clauses will never be reached for $q$ since only the first applicable clause in a decision list fires).
- If $B_{spec}$ equals *false*, we discard $C_q$ and continue by processing the next clause in $D$ (line 9).
- Otherwise, we assert a clause of the form
  `cpd_graduates(s1,[yes:0.2,no:0.8]) :- B`<sub>spec</sub>`, !.`
  (line 11), and we continue by processing the next clause in $D$ (line 12). □

The function SPECIALIZE_BODY (Figure 2) performs three steps which we explain in the next section. For comprehensibility we already give a simple example.

*Example 3.* Let the body to be specialized be `grade(s1,_C,a)`. First we ground the free variable $C$ (line 1 of SPECIALIZE_BODY), yielding a disjunction $B_1$, namely `grade(s1,c1,a) ; ... ; grade(s1,cn,a)`. Then we specialize each of the literals in $B_1$ with respect to the evidence (line 2). Consider the first literal, `grade(s1,c1,a)`. If we have evidence that $s1$ obtained grade 'a' for course $c1$ then we replace the literal by *true*, if we have different evidence we replace it by *false*, if we have no evidence we leave it unchanged. Doing this for each literal yields a disjunction $B_2$. Finally, we simplify $B_2$ using logical propagation rules (e.g. a disjunction is true if one if its disjuncts is true), yielding $B_3$ (line 3). □

There is one exception to the approach illustrated above. If there is no evidence at all about $B$, then $B_3$ would be identical to $B_1$: we would ground without specializing and simplifying, resulting in code explosion. In experiments we found that in such cases it is better for efficiency if we do not ground at all.[1] This special case is taken care of by lines 4 and 5 of SPECIALIZE_BODY.

### 5.2 Inner Loops of the Specialization Algorithm

We now further explain the three different steps in the function SPECIALIZE_BODY (namely GROUND_BODY, SPECIALIZE_LITERALS and SIMPLIFY_BODY).

---

[1] Leuschel and Bruynooghe [6] have similar observations about code explosion.

```
function SPECIALIZE_LITERAL(L, U, O, o)
1  if L is a positive (non-negated) literal
2     if L has an associated RV X ∈ O ∪ U
3        if X ∈ O   // there is evidence about X
4           if L is consistent with the evidence in o
5              return true
6           else return false
7        else return L   // no evidence, so no specialization
8     else return false   // non-existent RV
9  else // L is a negative (negated) literal
10    ...   // same as above but with role of true and false reversed
```

**Fig. 3.** Specializing a state literal $L$ with respect to the evidence.

**Step 1: Compact Grounding of Clause Bodies.** The first step takes as input the clause body to be specialized and partially grounds it. To be precise, we ground all free variables that occur in state literals (literals built from a state predicate).

First consider the case where the clause body consists of a *single state literal*. If the last argument of the literal is a free variable, we ground that variable with respect to the range of the state predicate. If other arguments of the literal are free variables, we ground them with respect to their population (for instance, for the variable $C$ in $grade(s1, C, a)$ this is the population of courses). The result is a disjunction of literals (since free variables in the body are existentially quantified). For instance, grounding `grade(s1,C,a)` gives a disjunction `grade(s1,c1,a) ; ... ; grade(s1,cn,a)`.

Now consider the case where the clause body is a *conjunction of literals*. We traverse the conjunction and whenever we encounter a state literal $L$ with free variables, we ground these variables as above (we ground them in $L$ and in all other literals that contain them). In principle, the result would be a disjunction of conjunctions. However, we try to represent the grounding as compactly as possible by means of a nested formula. For instance, we do not ground the conjunction `p,q(X)` as `(p,q(x1) ; ... ; p,q(xn))` but as `p,(q(x1);...;q(xn))`.[2]

**Step 2: Specialization of Individual Literals.** The previous grounding step paves the way for the actual specialization with respect to the evidence. We do this by traversing the nested formula constructed in the previous step. For each state literal $L$ that we encounter we perform specialization. Because of the previous grounding step, $L$ is guaranteed to be ground. Hence specializing $L$ with respect to the evidence is straightforward. The function that performs the specialization is shown in Figure 3.

---

[2] To obtain this nested formula, we recursively decompose the conjunction into independent components before grounding it. We do this decomposition in the same way as Santos Costa et al. [14] and Struyf [15, Ch.3] in their "once-transformation".

```
function SIMPLIFY_BODY(F)                5  else if F is a conjunction
1  if F is a disjunction                 6     for each conjunct F_i of F
2     for each disjunct F_i of F         7        S_i =SIMPLIFY_BODY(F_i)
3        S_i =SIMPLIFY_BODY(F_i)         8     return SIMPLIFY_CONJ(S_1 ∧ ... ∧ S_n)
4     return SIMPLIFY_DISJ(S_1 ∨ ... ∨ S_n)  9  else return F   // F is a single literal
```

**Fig. 4.** Simplifying a specialized formula $F$.

**Step 3: Simplification of Specialized Formulas.** The previous step does not change the structure of the nested formula that it operates on, but only the literals in the formula: some literals are replaced by $true$, some by $false$, the others are left unchanged. In a final step we simplify the resulting formula.

We traverse the given formula as shown in Figure 4. For a disjunction we first recursively simplify each of the disjuncts (line 3). Then we simplify the disjunction itself using two simple logical propagation rules: 1) if a disjunct is $true$, the entire disjunction is $true$; 2) disjuncts that are $false$ can be dropped (line 4). We deal with conjunctions in a similar way.

**Specialization of Meta-Predicates.** So far we considered only clause bodies that do not contain meta-predicates such as *findall*. While our specialization algorithm does not support all possible uses of meta-predicates, it does support several meta-predicate usage patterns that often occur in real-world probabilistic logical models.[3] One such case is that of a 'count *findall*', i.e. a *findall* whose purpose is to count the number of solutions to a query. To deal with this, our algorithm uses the same three steps as before: grounding, specialization, simplification. Let us explain this with a somewhat extreme but illustrative example.

*Example 4.* Consider the following clause body.

```
findall(C,grade(s1,C,a),L), length(L,N), N<2
```

First we ground the second argument of the *findall* as follows (the constant `d` denotes an auxiliary dummy element).

```
findall(d,( grade(s1,c1,a) ; ... ; grade(s1,c5,a) ),L),
                                              length(L,N), N<2
```

Next we specialize each of the *grade*/3 literals inside the *findall* (in the same way as we specialize other state literals). Suppose that we obtain the following.

```
findall(d,( grade(s1,c1,a) ; true ; grade(s1,c3,a) ; false ; true )
                                     ,L), length(L,N), N<2
```

---

[3] When our specialization algorithm encounters a meta-predicate usage pattern that is not supported, it simply leaves that part of the clause unchanged (unspecialized).

Next we simplify this conjunction. We can safely remove the *false* disjunct in the *findall*. We can also remove the two *true* disjuncts if we take them into account in the computation of the count `N`.

```
findall(d,( grade(s1,c1,a) ; grade(s1,c3,a) ),L), length(L,M),
                                                 N is M+2, N<2
```

Using simple arithmetic, this conjunction is further simplified to the following.

```
findall(d,( grade(s1,c1,a) ; grade(s1,c3,a) ),L), length(L,M), M<0
```

Since $M$ is at least 0, the entire conjunction is further simplified to *false*. □

### 5.3 Discussion.

From the perspective of efficiency of the specialization process, our specialization algorithm is clearly not optimal: the specialization time can easily be reduced, for instance by merging the three different steps. However, in experiments (Section 6.2) we observed that the specialization time is negligible as compared to the runtime of sampling with the specialized decision lists. Hence, we keep our specialization algorithm as simple as possible, rather than complicating it in order to reduce specialization time. This also makes it easier to see that specialization indeed preserves the semantics of the CPD-predicates, and hence that sampling produces the same samples as without specialization.

Note that the above remark is about the optimality of the specialization *process*, not of its *output*. As far as we can judge, the output (the specialized decision lists) is close to the optimal that can be achieved by specialization.

## 6 Experiments

We now experimentally analyze the influence of specialization on the efficiency of sampling. As a case study we consider the *Gibbs sampling* algorithm [1,13]. For pseudocode of this algorithm, see [4].

### 6.1 Setup

We use three real-world datasets: IMDB, UWCSE and WebKB. These datasets are common benchmarks in the area of probabilistic logical models [3]. In previous work we have applied machine learning algorithms to these datasets [5]. For each dataset, we converted the learned model to a parameterized Bayesian network. Table 1 gives some statistics about the models and the data (see Fierens et al. [5] for more information).

We use two inference scenarios, corresponding to the two scenarios of Section 4.3.

**Table 1.** Statistics about the data (number of RVs and parRVs) and the models (number of clauses in the decision lists

| Dataset | parRVs | RVs | Clauses |
|---------|--------|------|---------|
| IMDB    | 7      | 2852 | 13      |
| UWCSE   | 10     | 9607 | 32      |
| WebKB   | 5      | 78132| 12      |

- The first scenario is *classification*: there is one parRV that is the class, all RVs associated to that parRV are unobserved and need to be predicted, all other RVs are observed. For each dataset we do multiple experiments, each time with a different parRV as the class.[4]
- The second scenario is *missing data*: a random fraction $f$ of all RVs is unobserved ('missing'), the others are observed. We use several values of $f$, ranging from 5% to 50%. For each value we repeat each experiment 5 times with different unobserved RVs and we report the mean and standard deviation of the results across these 5 repetitions.

We measure the time to draw 10000 samples with the Gibbs sampling algorithm. Since our main goal is to investigate the *relative* efficiency of the different settings (with specialization versus without specialization), the choice of the number of samples does not heavily influence our conclusions. We report runtimes in minutes. The *runtime without specialization* is the runtime of sampling with parameterized CPDs that have not been grounded or specialized. The *runtime with specialization* is the sum of the specialization time and the runtime of sampling with specialized CPDs. Recall that both settings (with/without specialization) produce exactly the same sequence of samples.

### 6.2 Results

**Main results.** The results for the *missing data scenario* are shown in Figure 5. The results show that using specialization always yields a speedup. The magnitude of the speedup of course depends strongly on the amount of evidence. On WebKB, the dataset that is by far the most computationally demanding, we get a speedup of an order of magnitude when there are 5% unobserved RVs. On the smaller datasets (IMDB and UWCSE), the speedups are more modest.

The results for the *classification scenario* are shown in Table 2. For 4 of the 7 tasks, specialization yields significant speedups of a factor 4.3 to 6.5. For the other tasks, the speedup is small to negligible ($\leq 1.5$); these are mostly cases where the state predicate that forms the computational bottleneck (e.g. because it occurs inside a findall) is unobserved and hence cannot be specialized on.

---

[4] For more than half of all parRVs, predicting them is trivial (it can be done exactly in an efficient way, so there is no need for sampling). We exclude such parRVs as the class. For the WebKB dataset, we also exclude the parRV 'prof' as the class since this experiment timed-out.

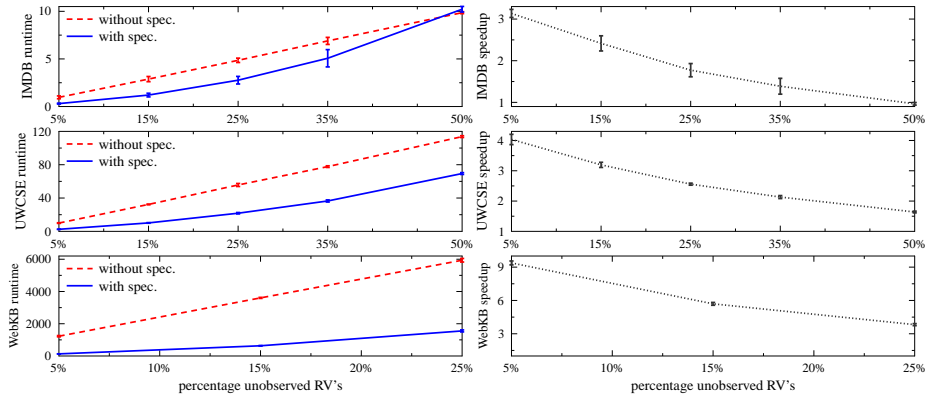

**Fig. 5.** Results for the missing data scenario. Left: runtime without specialization and with specialization. Right: speedup factor achieved due to specialization.

**Table 2.** Results for the classification scenario: runtime without specialization, runtime with specialization, and speedup factor achieved due to specialization.

| Data/Class | No spec. | Spec. | Speedup | Data/Class | No spec. | Spec. | Speedup |
|---|---|---|---|---|---|---|---|
| IMDB/acts | 16.1 | 14.9 | 1.08 | UWCSE/phase | 12.2 | 2.1 | 5.87 |
| IMDB/directs | 2.6 | 1.7 | 1.51 | UWCSE/teaches | 71.8 | 15.8 | 4.55 |
| UWCSE/advisedby | 75.1 | 17.4 | 4.31 | WebKB/hasproject | 2628 | 406 | 6.48 |
| UWCSE/coauthor | 10.9 | 10.4 | 1.05 | | | | |

**Influence of the data-size.** In the above results for both scenarios, the speedups are the lowest on the smallest dataset (IMDB) and the highest on the largest one (WebKB). This suggests a correlation between the speedup and the data-size (number of RVs). To investigate this we performed additional experiments in which we varied the data-size. Figure 6 shows the results for the missing data scenario with 15% unobserved RVs (results for the other settings are very similar). The trend in the speedup is clear: the larger the dataset, the higher the speedup. This is positive: speedups are more necessary on large datasets than on small ones.

**Efficiency.** The runtime with specialization is defined as the specialization time ($t_{spec}$) plus the runtime for sampling with specialized CPDs ($t_{sample}$). We also measured the fraction of time spent on specialization: $t_{spec}/(t_{spec} + t_{sample})$. This fraction is typically very low, on average 2.3%. This shows that (as argued in Section 5.2) there is no need to make the specialization process itself faster.

## 7 Conclusions

We considered the task of performing probabilistic inference with probabilistic logical models. We used the framework of parameterized Bayesian networks and showed how to implement sampling-based inference algorithms in Prolog. We argued that logic program specialization is suited to make sampling more efficient, which can in turn make the results more accurate. We developed a specialization algorithm and experimentally investigated the influence of specialization on the efficiency of Gibbs sampling. We found that specialization yields speedups of up to an order of magnitude and that these speedups grow with the data-size.

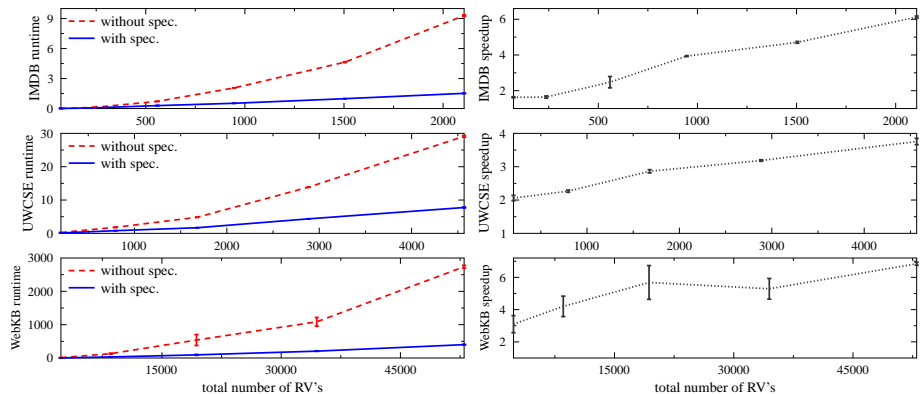

**Fig. 6.** Influence of the data-size (total number of RVs) for the missing data scenario with 15% unobserved RVs.

**Acknowledgments.** This research is supported by Research Foundation Flanders (FWO Vlaanderen), GOA/08/008 'Probabilistic Logic Learning' and Research Fund K.U.Leuven.

## References

1. B. Bidyuk and R. Dechter. Cutset sampling for Bayesian networks. *Journal of Artificial Intelligence Research*, 28:1–48, 2007.
2. C. M. Bishop. *Pattern Recognition and Machine Learning*. Springer, 2006.
3. L. De Raedt, P. Frasconi, K. Kersting, and S. Muggleton. *Probabilistic Inductive Logic Programming*. Springer, 2008.
4. D. Fierens. Improving the efficiency of gibbs sampling for probabilistic logical models by means of program specialization. In *Technical Communications of the 26th International Conference on Logic Programming (ICLP 2010)*, volume 7 of *Leibniz International Proceedings in Informatics (LIPIcs)*, pages 74–83, Dagstuhl, Germany, July 2010. Schloss Dagstuhl–Leibniz-Zentrum fuer Informatik.
5. D. Fierens, J. Ramon, M. Bruynooghe, and H. Blockeel. Learning directed probabilistic logical models: Ordering-search versus structure-search. *Annals of Mathematics and Artificial Intelligence*, 54(1):99–133, 2008.
6. M. Leuschel and M. Bruynooghe. Logic program specialisation through partial deduction: Control issues. *Theory and Practice of Logic Programming*, 2(4-5):461–515, 2002.

7. M. Leuschel, S. Craig, M. Bruynooghe, and W. Vanhoof. Specialising interpreters using offline partial deduction. In *Program Development in Computational Logic*, volume 3094 of *Lecture Notes in Computer Science*, pages 340–375. Springer, 2004.
8. X.-L. Li and Z.-H. Zhou. Structure learning of probabilistic relational models from incomplete relational data. In *Proceedings of the 18th European Conference on Machine Learning (ECML 2007)*, volume 4701 of *Lecture Notes in Computer Science*, pages 214–225. Springer, 2007.
9. R. Neapolitan. *Learning Bayesian Networks*. Prentice Hall, New Jersey, 2003.
10. A. Pettorossi and M. Proietti. Transformation of logic programs: Foundations and techniques. *Journal of Logic Programming*, 19-20:261–320, 1994.
11. D. Poole. First-order probabilistic inference. In *Proceedings of the 17th International Joint Conference on Artificial Intelligence (IJCAI 1997)*, pages 985–991. Morgan Kaufmann, 2003.
12. H. Poon and P. Domingos. Sound and efficient inference with probabilistic and deterministic dependencies. In *Proceedings of the 21st National Conference on Artificial Intelligence (AAAI 2006)*, pages 214–225. AAAI Press, 2006.
13. V. Santos Costa. On the implementation of the CLP(BN) language. In *Proceedings of the 12th International Symposium on Practical Aspects of Declarative Languages (PADL 2010)*, volume 5937 of *Lecture Notes in Artificial Intelligence*, pages 234–248. Springer, 2010.
14. V. Santos Costa, A. Srinivasan, R. Camacho, H. Blockeel, B. Demoen, G. Janssens, J. Struyf, H. Vandecasteele, and W. Van Laer. Query transformations for improving the efficiency of ILP systems. *Journal of Machine Learning Research*, 4:465–491, 2003.
15. J. Struyf. *Improving the efficiency of inductive logic programming in the context of relational data mining*. PhD thesis, Department of Computer Science, Katholieke Universiteit Leuven, December 2004.